\pdfoutput=1
\documentclass[11pt]{article}

\usepackage[]{ACL2023}

\usepackage{times}
\usepackage{latexsym}
\usepackage{booktabs}
\usepackage{svg}

\usepackage[T1]{fontenc}

\usepackage[utf8]{inputenc}

\usepackage{microtype}

\usepackage{inconsolata}

\usepackage{bm}

\usepackage{amsfonts}
\usepackage{amssymb}
\usepackage{amsmath}
\usepackage{makecell}
\usepackage{colortbl}

\usepackage{multirow}

\DeclareMathOperator*{\opMean}{mean}

\DeclareMathOperator*{\opAlign}{alignment}

\definecolor{darkgreen}{HTML}{00B050}
\definecolor{lightgreen}{HTML}{92D050}

\usepackage{xspace}

\usepackage{enumitem}
\usepackage{calc}

\newcommand{\alignmodel}[0]{{\textsc{AlignScore}}\xspace}
\title{\alignmodel : Evaluating Factual Consistency \\
with A Unified Alignment Function}

\author{Yuheng Zha\quad Yichi Yang \quad Ruichen Li\quad Zhiting Hu \\
UC San Diego \\
  \texttt{\{yzha, yiy067, rul014, zhh019\}@ucsd.edu} \\
}

\begin{document}
\maketitle
\begin{abstract}
Many text generation applications require the generated text to be factually consistent with input information. Automatic evaluation of factual consistency is challenging. Previous work has developed various metrics that often depend on \emph{specific} functions, such as natural language inference (NLI) or question answering (QA), trained on limited data. Those metrics thus can hardly assess diverse factual inconsistencies (e.g., contradictions, hallucinations) that occur in varying inputs/outputs (e.g., sentences, documents) from different tasks. In this paper, we propose \alignmodel, a new holistic metric that applies to a variety of factual inconsistency scenarios as above. \alignmodel is based on a \emph{general} function of {\it information alignment} between two arbitrary text pieces. Crucially, we develop a unified training framework of the alignment function by integrating a large diversity of data sources, resulting in 4.7M training examples from 7 well-established tasks (NLI, QA, paraphrasing, fact verification, information retrieval, semantic similarity, and summarization). We conduct extensive experiments on large-scale benchmarks including 22 evaluation datasets, where 19 of the datasets were never seen in the alignment training. \alignmodel achieves substantial improvement over a wide range of previous metrics. Moreover, \alignmodel (355M parameters) matches or even outperforms metrics based on ChatGPT and GPT-4 that are orders of magnitude larger.\footnote{Our code is available at \url{https://github.com/yuh-zha/AlignScore}.}
\end{abstract}

\section{Introduction}
Recent systems for natural language generation, such as summarization and dialogue systems, can produce fluent and coherent text. However, studies show the generated text can often contain factual consistency errors, such as contradictions with input information or hallucinations irrelevant to the context
\citep{cao2018faithful,kryscinski-etal-2019-neural,nie-etal-2019-simple,tan2020summarizing,maynez-etal-2020-faithfulness,deng-etal-2021-compression}.



It is thus crucial to develop automatic metrics that evaluate factual consistency of a \emph{claim} (e.g., generated text) with regard to a \emph{context} (e.g., model input). The evaluation, however, has long been a challenge. Recent work has devised various metrics based on specific pretrained functions, such as natural language inference (NLI) \citep{honovich-etal-2022-true,mishra-etal-2021-looking,kryscinski-etal-2020-evaluating,utama-etal-2022-falsesum,laban-etal-2022-summac} and question answering (QA) \citep{durmus-etal-2020-feqa,fabbri-etal-2022-qafacteval,honovich-etal-2021-q2,fabbri-etal-2022-qafacteval}. Specifically, an NLI-based metric measures if the claim is entailed by the context; while a QA-based metric first creates (question, answer) pairs from the claim and then checks if answering the questions with a QA model conditioning on the context will lead to the same answers. 

However, by relying on specific functions trained with only narrow data (i.e., NLI or QA datasets), previous metrics have limited generalizability and fail to apply to diverse evaluation scenarios, including different types of factual consistency errors and varying lengths and characteristics of contexts/claims from different tasks and domains. For instance, a metric trained exclusively with NLI data of sentences in a certain domain tends to have difficulty in evaluating summaries of long documents in a different domain \citep{mishra-etal-2021-looking, laban-etal-2022-summac}.
The limitations motivate a more holistic metric that develops a general understanding of factual consistency and generalizes to diverse evaluation scenarios.

In this paper, we propose \alignmodel, a new general factual consistency metric based on a unified text-to-text information alignment function. In particular, we unify a wide range of data sources, and use the massive diverse data to train a general information alignment model that estimates an alignment score given two arbitrary text pieces. More specifically, we reformat and aggregate 15 datasets from 7 popular language tasks, including NLI, QA, paraphrasing, fact verification, information retrieval, semantic similarity, and summarization. This results in a total of 4.7M training examples with diverse characteristics, and yields an alignment function with great generalizability. We then build \alignmodel using the alignment function as a building block. In particular, to handle long 
text and accommodate the different roles of context and claim, we develop a splitting strategy that breaks a context into coarse-grained chunks and a claim into fine-grained sentences. Aggregating the alignment scores between context-chunks and claim-sentences leads to the final factual consistency score. 



In our experiments, we build \alignmodel by finetuning the lightweight RoBERTa models (125M and 355M) for alignment.
We evaluate \alignmodel on the latest large-scale evaluation benchmarks, including SummaC \citep{laban-etal-2022-summac}, TRUE \citep{honovich-etal-2022-true-evaluating}, and other testbeds, which contain a total of 22 challenging evaluation datasets. 
Our approach substantially outperforms previous state-of-the-art metrics in terms of different quality measures. Notably, our metric (355M) is on par with, and sometimes even much better than latest metrics based on orders-of-magnitude larger language models (e.g., ChatGPT and GPT-4). 
In particular, \alignmodel shows strong generalizability on the 19 zero-shot datasets that were never seen during the alignment function training. We also conduct extensive ablation studies to demonstrate the effectiveness of the context splitting strategy and other modeling choices.


\section{Related Work}

\paragraph{Factual Consistency Metrics}
Traditionally, generative systems are evaluated using n-gram based metrics \citep{papineni-etal-2002-bleu,lin-2004-rouge,banerjee-lavie-2005-meteor,popovic-2015-chrf}. Recently, factual consistency metrics are often use task-specific language understanding capabilities, such as NLI and QA. To improve performance when evaluating generative tasks with long texts, NLI-based metrics adopt training sets with long premises \citep{honovich-etal-2022-true,mishra-etal-2021-looking}, use large synthetic datasets \citep{kryscinski-etal-2020-evaluating, utama-etal-2022-falsesum}, or use sentence level evaluation \citep{laban-etal-2022-summac}. A separate line of research formulates factual consistency evaluation as QA \citep{durmus-etal-2020-feqa, fabbri-etal-2022-qafacteval,honovich-etal-2021-q2,fabbri-etal-2022-qafacteval}. 
Other consistency evaluation methods that use pretrained language models (LMs) include embedding matching \citep{zhang2020bertscore,deng-etal-2021-compression}, 
finetuning LMs to directly regress human evaluation scores \citep{sellam-etal-2020-bleurt}, and using LMs to score candidates based on weighted log probability \citep{yuan2021bartscore,liu2022maskeval}. 
CTC \citep{deng-etal-2021-compression} develops a suite of text generation evaluation metrics based on the similar concept of alignment. Yet we define alignment in a more general way to enable integration of diverse training data, and deliver \alignmodel as a more effective metric focusing on factual consistency.
Concurrent work proposes to combine large language models (LLMs) with prompting to evaluate different aspects of generated text, including factual consistency \citep{fu-etal-2023-gptscore,liu-etal-2023-g-eval,gao-etal-2023-human}. Our proposed \alignmodel shows stronger performance with a much smaller model size.


\paragraph{Unified Training}
Recent work converts related but different tasks into the same input-output format to train unified models.
\citet{raffel2020exploring} propose to unify text generation tasks into a text-to-text conditional generation problem. \citet{sanh2022multitask} further show that the text-to-text generation framework, combined with natural language prompting, improves zero-shot task generalization to unseen tasks. 
\citet{Zhong2022TowardsAU} develop a unified automatic evaluation metric by framing different aspects of NLG evaluation as a Boolean Question Answering problem. 
Recent studies also present task unification as an effective approach to improve model performance and generalizability in multi-modal tasks \citep{xie-etal-2022-unifiedskg,zhang2021vinvl,wang2022ofa}.

\begin{figure*}[t]
    \centering
    \includegraphics[width=\textwidth]{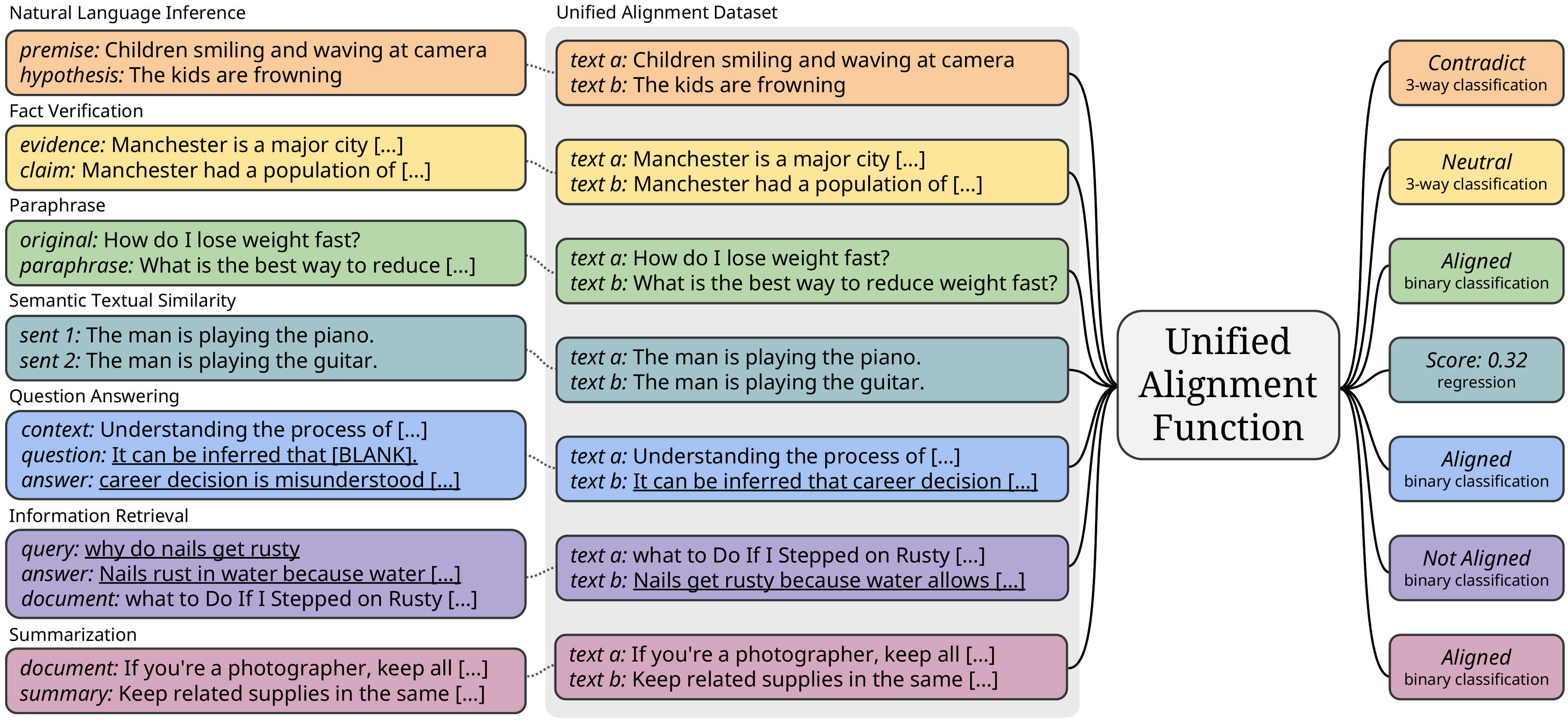}
    \caption{A diagram illustrating the information alignment problem and how we unify various tasks into the alignment task. We convert each sample in the tasks we consider into a text pair \((\boldsymbol{a}, \boldsymbol{b})\), and the alignment function predicts a label \(y\) characterizing the level of alignment. The \underline{underlined} text indicates items in the original dataset (e.g., question and answer in a QA dataset) are combined to form part of the text pair in the alignment dataset.}
    \label{fig:alignment-task}
\end{figure*}

\begin{table*}[ht]
\centering 
\resizebox{\textwidth}{!}{%
\begin{tabular}{@{}lllrrr@{}}
\toprule
NLP Task                                    & Dataset                                           & Training Task         & \multicolumn{2}{l}{Avg. Word Count}                     & \multicolumn{1}{l}{Sample Count} \\ \cmidrule(lr){4-5}
                                            &                                                   &                       & \multicolumn{1}{l}{Context} & \multicolumn{1}{l}{Claim} & \multicolumn{1}{l}{}             \\ \midrule
\textit{NLI}                                & SNLI \citep{bowman-etal-2015-large}               & 3-way classification  & 13                          & 7                         & 550k                             \\
\textit{}                                   & MultiNLI \citep{williams2018broad}                & 3-way classification  & 20                          & 10                        & 393k                             \\
\textit{}                                   & Adversarial NLI \citep{nie-etal-2020-adversarial} & 3-way classification  & 54                          & 10                        & 163k                             \\
\textit{}                                   & DocNLI \citep{yin-etal-2021-docnli}               & binary classification & 285                         & 43                        & 942k                             \\ \midrule
\multirow{2}{*}{\textit{Fact Verification}} & NLI-style FEVER \citep{DBLP:conf/aaai/NieCB19}          & 3-way classification  & 50                          & 8                         & 208k                             \\
                                            & Vitamin C \citep{schuster-etal-2021-get}          & 3-way classification  & 25                          & 11                        & 371k                             \\ \midrule
\multirow{3}{*}{\textit{Paraphrase}}        & QQP \citep{csernai-qqp}                           & binary classification & 11                          & 11                        & 364k                             \\
                                            & PAWS \citep{zhang-etal-2019-paws}                 & binary classification & 18                          & 18                        & 707k                             \\
                                            & WikiText-103* \citep{merity2017pointer}           & binary classification & 22                          & 21                        & 8M                               \\ \midrule
\multirow{2}{*}{\textit{STS}}               & SICK \citep{marelli-etal-2014-sick}               & regression            & 10                          & 10                        & 4k                               \\
                                            & STS Benchmark \citep{cer2017semeval}              & regression            & 10                          & 10                        & 6k                               \\ \midrule
\multirow{2}{*}{\textit{QA}}                & SQuAD v2 \citep{rajpurkar-etal-2018-know}         & binary classification & 119                         & 11                        & 130k                             \\
                                            & RACE \citep{lai-etal-2017-race}                   & binary classification & 273                         & 14                        & 351k                             \\ \midrule
\textit{Information Retrieval}              & MS MARCO \citep{nguyen2016msmarco}                & binary classification & 56                          & 15                        & 5M                               \\ \midrule
\textit{Summarization}                      & WikiHow* \citep{koupaee2018wikihow}               & binary classification & 508                         & 46                        & 157k                             \\ \bottomrule
\end{tabular}
}
\caption{The training datasets of our alignment model. Datasets marked with a * (WikiText-103, WikiHow) are augmented with synthetic samples (see Appendix~\ref{sec:synthetic-data}). Note due to resource constraints, we only use at most 500k samples from each dataset to train the alignment model.}
\label{tab:training-data}
\end{table*}

\section{Methods}

We introduce the \alignmodel metric built on top of a unified alignment function. We first train the  alignment function by unifying a large diversity of data sources (Section~\ref{sec:alignment-task}). We then define \alignmodel by combining the alignment function with a new context/claim splitting and aggregation strategy (Section~\ref{sec:consistency-metric}).

\subsection{Unified Alignment Function}\label{sec:alignment-task}
Given two pieces of text \(\boldsymbol{a}\) and \(\boldsymbol{b}\), we consider \(\boldsymbol{b}\) to be aligned with \(\boldsymbol{a}\) if all information in \(\boldsymbol{b}\) is present in \(\boldsymbol{a}\) and does not contradict \(\boldsymbol{a}\). Conceptually, we model information alignment as a function  that maps the text pair \((\boldsymbol{a},\boldsymbol{b})\) to a label \(y\) that characterizes the level of alignment:
\begin{align}
    f:(\boldsymbol{a},\boldsymbol{b}) &\rightarrow y\;\text{.}\label{eq:alignment}
\end{align}

A holistic and generalizable alignment function must account for all types of consistency errors, domains, and data distributions. Therefore, in order to learn the alignment function, we want to adapt and aggregate diverse language tasks to form a unified alignment training corpus (Figure~\ref{fig:alignment-task}). In this work, we collect 15 datasets spanning 7 well-established tasks, including NLI, fact verification, paraphrase, semantic textual similarity, QA, information retrieval, and summarization. We present an overview of these datasets in Table~\ref{tab:training-data} and include more details in Section \ref{sec:training-data} and \ref{sec:synthetic-data} in the appendix.

The vast diversity of input/output formats across the above tasks poses significant challenge for unifying them into a uniform alignment training corpus.
To unify input formats, we convert each sample into a text pair \((\boldsymbol{a}, \boldsymbol{b})\). For tasks that do not cleanly fit into the text pair format, such as QA (where each sample contains a question, an answer, and a context) and information retrieval (where each sample contains a query, an answer, and a supporting document), we use a sequence-to-sequence model \citep{song-qa2d} to convert the question answer pair into a single declarative sentence (\underline{underlined} items in Figure~\ref{fig:alignment-task}; See Section~\ref{sec:qa-samples} for examples).

To unify output formats, while it is possible to transform all tasks into binary classification, instead we convert them into a set of related alignment problems to preserve as much information as possible from the original datasets (Figure~\ref{fig:alignment-task}). Specifically, we devise 3 options for the alignment label $y$:
\begin{align*}
    y_{\text{bin}} &\in \{\textsc{aligned}, \textsc{not-aligned}\}, \\
    y_{\text{3way}} &\in \{\textsc{aligned}, \textsc{contradict}, \textsc{neutral}\}, \\
    y_{\text{reg}} &\in [0,1].
\end{align*}
More concretely, for tasks that come with discrete labels, depending on their setup, the alignment function predicts either the binary classification label \(y_{\text{bin}}\) (paraphrase, QA, information retrieval, and summarization) or the 3-way classification label \(y_{\text{3way}}\) (NLI, and fact verification); for tasks with continuous labels (semantic textual similarity), the alignment function predicts the regression label \(y_{\text{reg}}\). Here a higher \(y_{\text{reg}}\) indicates that more information in \(\boldsymbol{b}\) is supported by \(\boldsymbol{a}\).

We build the alignment model consisting of a language model (e.g., RoBERTa; \citealp{liu2019roberta}) and 3 individual linear layers as the 3-way classification (\(y_{\text{3way}}\)), binary classification (\(y_{\text{bin}}\)), and regression (\(y_{\text{reg}}\)) heads. First, we feed into the language model the concatenation of the text pair \((\boldsymbol{a}, \boldsymbol{b})\) and use the contextual embedding of the special begin-of-sentence token as the encoded representation, \(\boldsymbol{h}\). Then, the classification and regression heads map \(\boldsymbol{h}\) into an estimation of \(y_{\text{3way}}\), \(y_{\text{bin}}\), and \(y_{\text{reg}}\) through logistic regression and linear regression, respectively.
We use cross entropy loss for both 3-way and binary classification, and mean squared error loss for regression. The joint loss function is:
\begin{equation}
    \mathcal{L}_{\text{total}} = \lambda_1\mathcal{L}_{\text{3way}} + \lambda_2\mathcal{L}_{\text{bin}} + \lambda_3\mathcal{L}_{\text{reg}},
\label{eq:total-loss}
\end{equation}
where $\lambda_1, \lambda_2, \lambda_3$ are scalar weights. In our experiments, we set $\lambda_1=\lambda_2=\lambda_3=1$.

\begin{figure}[t]
    \centering
    \includegraphics[width=\columnwidth]{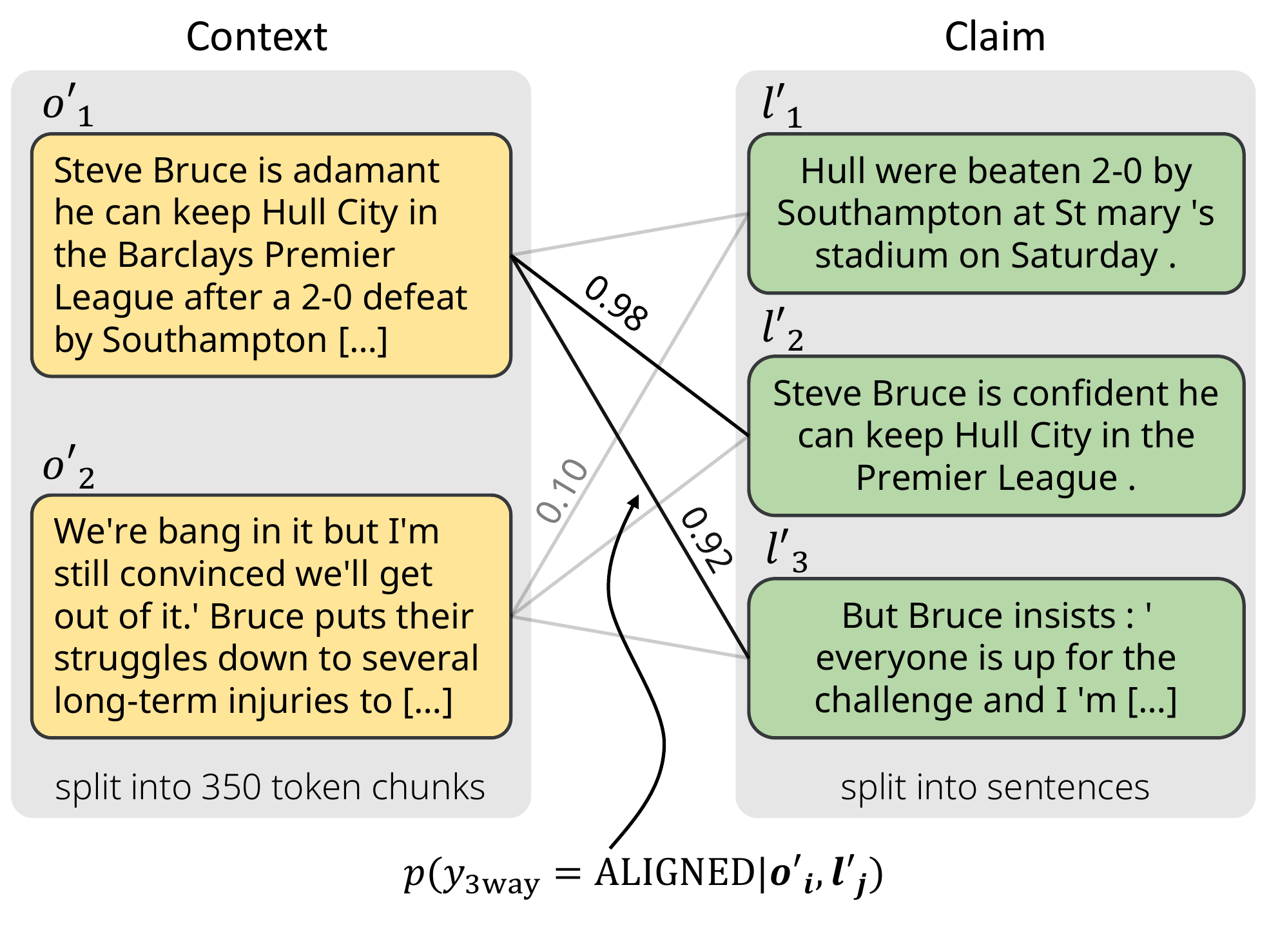}
    \vspace{-15pt}
    \caption{Illustration of \alignmodel. The \textit{context} is split into roughly 350-token chunks. Then, each sentence in the \textit{claim} is evaluated against the \textit{context} chunks using the alignment function. The highest alignment score of each \textit{claim} sentence is selected and then averaged to derive the factual consistency score.}
    \label{fig:splitting}
\end{figure}

\subsection{The \alignmodel Metric}\label{sec:consistency-metric}


As the definition of factual consistency is closely related to the information alignment problem, one naive way of building a factual consistency metric is simply using the alignment model to estimate the alignment score of the text pair (\textit{context}, \textit{claim}). However, this approach (also referred to as "document level evaluation"; \citealp{laban-etal-2022-summac}) has several drawbacks.

First, generative tasks often contain long inputs, especially long \textit{contexts}, that go beyond the input length limit of a language model (e.g., source documents in summarization tasks can easily exceed the 512-token limit of a RoBERTa model). Consequently, if long inputs are not explicitly handled \citep{kryscinski-etal-2020-evaluating,mishra-etal-2021-looking}, language-model-based metrics could silently drop important information because of truncation.

Second, information contained in a \textit{claim} often spreads across multiple sentences in the \textit{context}. To verify the factual consistency of a \textit{claim}, a metric needs access to long \textit{context} spans. Therefore, evaluating the \textit{claim} against individual \textit{context} sentences (as in previous sentence level evaluation; \citealp{laban-etal-2022-summac, amplayo2022smart}) can degrade metric performance as paragraph- and document-level semantic information is lost.

Third, humans typically assign consistency scores in a continuous spectrum that reflect the amount of consistency errors in the samples. Similarly, good metrics should produce fine-grained scores. Unfortunately, as classification tasks make up most of the training data (only semantic textual similarity datasets provide continuous labels), our alignment model tends to assign scores close to the two extremes, limiting its effectiveness if used directly as a factual consistency metric.

Conceptually, to resolve the first challenge, we need to split the \textit{context} into chunks such that when concatenated with a \textit{claim}, the resulting sequence does not exceed the input length limit. By picking a large enough chunk size, we allow the model to reason over longer \textit{context} spans, mitigating the second issue. Since sentences in a \textit{claim} tend to be self-contained statements, an effective way to make the metric produce more fine-grained scores is to evaluate \textit{claim} sentences independently of each other \cite{laban-etal-2022-summac}. Specifically, for each sentence in the \textit{claim} (green rectangles in Figure~\ref{fig:splitting}), we evaluate it against all \textit{context} chunks (yellow rectangles in Figure~\ref{fig:splitting}) using the alignment function. Then, we select the highest alignment score (lines labeled with numbers in Figure~\ref{fig:splitting}) for each \textit{claim} sentence. Intuitively, this step identifies the \textit{context} chunk that most strongly supports each \textit{claim} sentence, and the highest score reflects how well the \textit{claim} sentence is supported. Finally, we use the average value of all highest scores as the factual consistency score. This addresses the third challenge, as taking the average prevents a single inconsistent \textit{claim} sentence from dominating the final score. Alternatively, the average value of highest scores can be roughly interpreted as "the proportion of the \textit{claim} that are factually consistent with respect to the \textit{context}", which naturally leads to a more fine-grained metric. As we show in experiments, our novel chunk level evaluation method consistently outperforms document level (which risks truncation) and sentence level evaluation.

We formally define \alignmodel as:
\begin{multline}
    \alignmodel(\boldsymbol{o}, \boldsymbol{l}) \\
    = \opMean_j \max_i \opAlign(\boldsymbol{o'_i}, \boldsymbol{l'_j})\;\text{,} \label{eq:align-metric}
\end{multline}
where \(\boldsymbol{o}\) is the \textit{context}, \(\boldsymbol{l}\) is the \textit{claim}, \(\{\boldsymbol{o'_i}\}\) is the set of \textit{context} chunks, \(\{\boldsymbol{l'_j}\}\) is the set of \textit{claim} sentences, and \(\opAlign(\cdot)\) is the probability of the model predicting the \textsc{aligned} label in the 3-way classification setting. In practice, for RoBERTa models (that have an input length limit of 512 tokens) we split the \textit{context} into chunks at sentence boundaries such that each chunk contains roughly 350 tokens. We use the output of the 3-way classification head, our ablation studies reveal  that it performs better than the binary classification head and the regression head (Section~\ref{sec:ablation-task}).


\begin{table*}[htbp]
\centering \footnotesize
\begin{tabular}{@{}c|lcccccc|c@{}}
\toprule
Type                         & Metric      & CGS   & XSF                           & PolyTope & FactCC & SummEval & FRANK & \textbf{AVG} \\ \midrule
                           & FEQA        & 53.7 & 47.6                         & 54.3    & 47.9  & 48.8    & 37.2 & 48.3        \\
                           & QuestEval   & 60.4 & 63.6                         & 77.0    & 74.2  & 74.3    & 85.8 & 72.5        \\
\multirow{-3}{*}{QA}         & QAFactEval  & 83.4 & 66.1                         & 86.4    & 89.2  & 88.1    & 89.4 & 83.8        \\ \midrule
                           & ROUGE-1     & 69.7 & 64.5                         & 82.5    & 75.8  & 87.2    & 85.0 & 77.4        \\
                           & ROUGE-2     & 70.5 & 65.9                         & 83.7    & 76.0  & 87.2    & 85.3 & 78.1        \\
                           & ROUGE-L     & 70.2 & 62.9                         & 81.9    & 76.3  & 87.3    & 85.3 & 77.3        \\
                           & BLEU        & 71.8 & 55.8                         & 86.9    & 75.0  & 83.8    & 84.5 & 76.3        \\
                           & BERTScore   & 63.1 & 49.0                         & 85.3    & 70.9  & 79.6    & 84.9 & 72.1        \\
                           & NER-Overlap & 51.1 & 64.9                         & 72.1    & 49.8  & 56.6    & 68.1 & 60.4        \\
\multirow{-7}{*}{\begin{tabular}[c]{@{}c@{}}Similarity\\ Matching\end{tabular}} &
SimCSE &
56.2 &
62.2 &
75.2 &
59.0 &
77.2 &
74.8 &
67.4 \\ \midrule
Regression                   & BLEURT      & 60.8 & 64.7                         & 76.7    & 59.7  & 71.1    & 82.5 & 69.2        \\ \midrule
                           & MNLI        & 44.9 & 46.6                         & 45.0    & 48.3  & 43.5    & 59.3 & 47.9        \\
                           & DAE         & 52.4 & \cellcolor[HTML]{E0F1CB}76.7 & 72.8    & 54.2  & 66.1    & 78.9 & 66.8        \\
                           & SummaC-ZS   & 73.6 & 58.0                         & 87.5    & 83.7  & 85.8    & 85.3 & 79.0        \\
\multirow{-4}{*}{NLI} & SummaC-CONV & 67.2 & 70.3                         & 81.8    & 92.3  & 86.1    & 88.5 & 81.0        \\ \midrule
&
UniEval &
\cellcolor[HTML]{E0F1CB}84.7 &
65.5 &
\cellcolor[HTML]{92D050}93.4 &
89.9 &
86.3 &
88.0 &
84.6 \\
                           & CTC         & 76.5 & 65.9                         & 89.5    & 82.6  & 85.6    & 87.3 & 81.2        \\
                           & BARTScore   & 74.3 & 62.6                         & 91.7    & 82.3  & 85.9    & 88.5 & 80.9        \\
                           & FactCC      & 64.9 & 55.1                         & 78.5    & 72.7  & 71.8    & 69.8 & 68.8        \\
\multirow{-5}{*}{Misc}       & BLANC       & 54.1 & 53.5                         & 74.7    & 56.4  & 68.6    & 83.4 & 65.1        \\ \midrule
&
\textbf{\alignmodel-base} &
83.7 &
\cellcolor[HTML]{92D050}79.4 &
87.8 &
\cellcolor[HTML]{E0F1CB}93.3 &
\cellcolor[HTML]{E0F1CB}89.9 &
\cellcolor[HTML]{E0F1CB}90.5 &
\cellcolor[HTML]{E0F1CB}87.4 \\
\multirow{-2}{*}{Ours} &
\textbf{\alignmodel-large} &
\cellcolor[HTML]{92D050}86.4 &
75.8 &
\cellcolor[HTML]{E0F1CB}92.4 &
\cellcolor[HTML]{92D050}93.7 &
\cellcolor[HTML]{92D050}91.7 &
\cellcolor[HTML]{92D050}91.4 &
\cellcolor[HTML]{92D050}88.6 \\ \bottomrule
\end{tabular}
\caption{The AUC-ROC of different metrics on the SummaC benchmark. The last column (\textbf{AVG}) is the average performance of each metric. The \textcolor{darkgreen}{dark green} indicates the best metric on each dataset or on average. And the \textcolor{lightgreen}{light green} indicates the second best. CGS and XSF are abbreviations for CoGenSumm and XSumFaith, respectively.}
\label{tab:summac-auc}
\end{table*}
  
\section{Experiments}
In this section, we evaluate \alignmodel on a wide range of benchmarks and show it consistently outperforms existing metrics (Section~\ref{sec:implementation}-\ref{sec:results}). We also conduct extensive ablation study in Section~\ref{sec:ablation-task}.
\subsection{Implementation}\label{sec:implementation}
We use RoBERTa \citep{liu2019roberta} to implement the alignment model. 
We denote \alignmodel based on RoBERTa-base/large as \alignmodel-base/large.

We follow common practice \citep{liu2019roberta, devlin-etal-2019-bert} and train the model for 3 epochs with a batch size of 32 in all the experiments. Training samples are randomly sampled across the converted upstream NLP tasks. Due to resource constraints we only use the first 500k samples in each dataset for training, resulting in a total of 4.7 million training samples. Training details are listed in Appendix~\ref{sec:training-details}.

\subsection{Benchmarks}\label{sec:benchmarks}
\label{sec:exp-benchmarks}
Following \citet{deng-etal-2021-compression}, \citet{fabbri-etal-2022-qafacteval}, \citet{Zhong2022TowardsAU} and \citet{gabriel-etal-2021-go}, we evaluate factual consistency metrics using TRUE benchmark \citep{honovich-etal-2022-true} (consists of 11 datasets in diverse domains), SummaC benchmark \citep{laban-etal-2022-summac} (includes 6 large summarization datasets), and a set of other latest datasets including XSumFaith \citep{maynez-etal-2020-faithfulness}, SummEval \citep{fabbri-etal-2021-summeval}, QAGS-XSum \citep{wang-etal-2020-asking}, QAGS-CNNDM \citep{wang-etal-2020-asking}, FRANK \citep{pagnoni-etal-2021-understanding} and SamSum \citep{gliwa-etal-2019-samsum}.

SummaC benchmark standardizes the task of summary inconsistency detection by casting it as a binary classification problem. Following \citet{laban-etal-2022-summac}, we 1) tune the threshold of metrics on the validation sets, and then compute the balanced accuracy \citep{brodersen2010balanced} on the test sets, 2) report the AUC-ROC \citep{bradley1997use} of each metric.
TRUE benchmark covers summarization, dialogue, paraphrase and fact verification tasks. It also assigns binary labels to samples based on whether the entire \textit{claim} is factually consistent with the \textit{context}. We report AUC-ROC of each metric following \citet{honovich-etal-2022-true}. 
We also collect 6 popular factual consistency evaluation datasets, namely XSumFaith, SummEval, QAGS-XSum, QAGS-CNNDM, FRANK and SamSum. We compute instance-level Pearson, Spearman, and Kendall's tau correlation coefficients between metric scores and human annotated consistency scores. 

\begin{table*}[htbp]
\centering 
\resizebox{\textwidth}{!}{
\begin{tabular}{@{}c|lccccccccccc|cc@{}}
\toprule
Type &
  Metric &
  SE &
  PAWS &
  Q2 &
  VitC &
  FVR &
  FRK &
  DF &
  MNBM &
  Q-C &
  Q-X &
  BEGIN &
  \textbf{AVG} &
  \textbf{AVG-ZS} \\ \midrule
 &
  FEQA &
  49.5 &
  50.0 &
  53.2 &
  49.9 &
  51.1 &
  63.0 &
  50.5 &
  48.8 &
  50.1 &
  49.4 &
  53.0 &
  51.7 &
  52.2 \\
 &
  QuestEval &
  69.7 &
  69.0 &
  72.2 &
  66.6 &
  72.5 &
  84.0 &
  77.2 &
  64.8 &
  64.5 &
  55.2 &
  83.9 &
  70.9 &
  71.4 \\
\multirow{-3}{*}{QA} &
  QAFactEval &
  80.9 &
  86.1 &
  75.8 &
  73.6 &
  86.0 &
  88.5 &
  81.8 &
  67.3 &
  83.9 &
  76.1 &
  81.0 &
  80.1 &
  79.4 \\ \midrule
 &
  ROUGE-1 &
  80.4 &
  50.2 &
  59.7 &
  60.9 &
  57.8 &
  83.6 &
  65.3 &
  64.8 &
  77.3 &
  60.1 &
  84.6 &
  67.7 &
  72.0 \\
 &
  ROUGE-2 &
  79.4 &
  68.6 &
  61.4 &
  59.9 &
  55.5 &
  84.5 &
  67.7 &
  65.0 &
  78.4 &
  60.2 &
  82.8 &
  69.4 &
  72.4 \\
 &
  ROUGE-L &
  80.4 &
  75.9 &
  60.6 &
  59.7 &
  56.4 &
  83.6 &
  65.4 &
  62.8 &
  77.6 &
  59.3 &
  85.0 &
  69.7 &
  71.8 \\
 &
  BLEU &
  74.8 &
  71.3 &
  55.2 &
  56.1 &
  51.7 &
  84.1 &
  61.2 &
  56.7 &
  77.4 &
  54.7 &
  74.6 &
  65.2 &
  67.3 \\
 &
  BERTScore &
  72.3 &
  78.6 &
  70.2 &
  58.2 &
  54.2 &
  84.0 &
  68.6 &
  52.5 &
  70.6 &
  44.3 &
  86.4 &
  67.2 &
  68.6 \\
 &
  NER-Overlap &
  56.6 &
  51.7 &
  59.1 &
  57.8 &
  62.4 &
  65.5 &
  62.7 &
  68.4 &
  48.4 &
  63.6 &
  50.6 &
  58.8 &
  59.3 \\
\multirow{-7}{*}{\begin{tabular}[c]{@{}c@{}}Similarity\\ Matching\end{tabular}} &
  SimCSE &
  70.2 &
  69.2 &
  66.2 &
  63.8 &
  72.7 &
  72.9 &
  70.6 &
  64.6 &
  74.9 &
  56.5 &
  86.1 &
  69.8 &
  70.3 \\ \midrule
Regression &
  BLEURT &
  68.0 &
  68.4 &
  72.9 &
  61.8 &
  59.5 &
  81.6 &
  73.0 &
  65.5 &
  71.2 &
  56.2 &
  \cellcolor[HTML]{E0F1CB}86.6 &
  69.5 &
  71.9 \\ \midrule
 &
  MNLI &
  44.6 &
  81.3 &
  71.8 &
  80.2 &
  93.1 &
  57.2 &
  76.5 &
  59.1 &
  42.6 &
  50.1 &
  81.5 &
  67.1 &
  60.4 \\
 &
  DAE &
  60.3 &
  55.8 &
  57.7 &
  60.2 &
  77.8 &
  77.9 &
  54.7 &
  \cellcolor[HTML]{92D050}81.0 &
  56.9 &
  67.5 &
  69.4 &
  65.4 &
  65.7 \\
 &
  SummaC-ZS &
  77.6 &
  89.0 &
  \cellcolor[HTML]{92D050}81.8 &
  97.2 &
  92.8 &
  86.9 &
  \cellcolor[HTML]{92D050}87.1 &
  58.0 &
  76.0 &
  75.3 &
  83.2 &
  82.2 &
  78.2 \\
\multirow{-4}{*}{NLI} &
  SummaC-CONV &
  79.1 &
  88.2 &
  77.5 &
  97.5 &
  92.0 &
  89.0 &
  81.2 &
  67.2 &
  77.7 &
  76.0 &
  81.6 &
  82.5 &
  78.7 \\ \midrule
 &
  UniEval &
  \cellcolor[HTML]{E0F1CB}81.2 &
  80.1 &
  70.4 &
  79.1 &
  92.1 &
  88.1 &
  80.4 &
  66.8 &
  86.5 &
  76.7 &
  73.6 &
  79.5 &
  78.0 \\
 &
  CTC &
  79.8 &
  63.1 &
  66.8 &
  65.0 &
  72.5 &
  87.1 &
  63.7 &
  65.0 &
  77.3 &
  67.7 &
  72.0 &
  70.9 &
  72.4 \\
 &
  BARTScore &
  78.9 &
  77.1 &
  65.1 &
  64.2 &
  66.1 &
  87.8 &
  60.8 &
  63.5 &
  83.9 &
  60.2 &
  \cellcolor[HTML]{92D050}86.7 &
  72.2 &
  73.4 \\
 &
  FactCC &
  68.6 &
  53.4 &
  59.3 &
  54.7 &
  58.7 &
  70.7 &
  55.0 &
  56.1 &
  70.1 &
  64.4 &
  57.6 &
  60.8 &
  62.7 \\
\multirow{-5}{*}{Misc} &
  BLANC &
  63.3 &
  56.0 &
  62.9 &
  55.7 &
  53.6 &
  82.1 &
  63.8 &
  54.2 &
  60.9 &
  50.9 &
  73.7 &
  61.6 &
  64.0 \\ \midrule
 &
  \textbf{\alignmodel-base} &
  80.8 &
  \cellcolor[HTML]{E0F1CB}97.3 &
  76.1 &
  \cellcolor[HTML]{E0F1CB}97.8 &
  \cellcolor[HTML]{E0F1CB}94.6 &
  \cellcolor[HTML]{E0F1CB}90.0 &
  83.1 &
  \cellcolor[HTML]{E0F1CB}79.9 &
  \cellcolor[HTML]{E0F1CB}87.7 &
  \cellcolor[HTML]{E0F1CB}79.6 &
  82.4 &
  \cellcolor[HTML]{E0F1CB}86.3 &
  \cellcolor[HTML]{E0F1CB}82.5 \\
\multirow{-2}{*}{Ours} &
  \textbf{\alignmodel-large} &
  \cellcolor[HTML]{92D050}82.9 &
  \cellcolor[HTML]{92D050}98.4 &
  \cellcolor[HTML]{E0F1CB}78.6 &
  \cellcolor[HTML]{92D050}98.3 &
  \cellcolor[HTML]{92D050}94.9 &
  \cellcolor[HTML]{92D050}92.1 &
  \cellcolor[HTML]{E0F1CB}85.1 &
  76.1 &
  \cellcolor[HTML]{92D050}89.5 &
  \cellcolor[HTML]{92D050}83.5 &
  82.7 &
  \cellcolor[HTML]{92D050}87.4 &
  \cellcolor[HTML]{92D050}83.8 \\ \bottomrule
\end{tabular}
}
\caption{The AUC-ROC of various metrics reported on TRUE benchmark. We compute both the overall average performance in the \textbf{AVG} column and the average without VitaminC, FEVER and PAWS datasets in the \textbf{AVG-ZS} column. The color format is the same as in Table \ref{tab:summac-auc}. The full names of the datasets are listed in Table \ref{tab:true-abbr}.}
\label{tab:true-benchmark}
\end{table*}

\subsection{Baselines}\label{sec:baselines}
We compare \alignmodel with state-of-the-art metrics, which we categorize into question answering (QA), similarity matching, regression, NLI, and miscellaneous. We use open-source code and models released by authors. Additionally, we also compare with latest LLM-based metrics.

\textbf{QA Based Metrics} adapt question generation (QG) and question answering (QA) models to automatically evaluate factual consistency.
We include the latest QAFactEval \citep{fabbri-etal-2022-qafacteval}, QuestEval \citep{scialom-etal-2021-questeval}, and FEQA \citep{durmus-etal-2020-feqa} as our baselines. 

\textbf{Similarity Matching Based Metrics} vary in their granularity and matching functions. We report BLEU \citep{papineni-etal-2002-bleu} and ROUGE-1/2/L \citep{lin-2004-rouge}, which compute token-level string matching scores. We also include the named-entity level metric NER-Overlap introduced in \citet{laban-etal-2022-summac}.
BERTScore \citep{zhang2020bertscore} uses token-level embedding to compute scores, for which we use the best variant (\texttt{microsoft/deberta-xlarge-mnli}) recommended by the authors\footnote{\url{https://github.com/Tiiiger/bert_score}}. We also use SimCSE \citep{gao-etal-2021-simcse} as sentence-level embedding matching function, with the best released model \texttt{sup-simcse-roberta-large\footnote{\url{https://github.com/princeton-nlp/SimCSE}}}.

\textbf{Regression Based Metrics} learn to estimate ground truth scores directly. We use BLEURT \citep{sellam-etal-2020-bleurt} 
with its recommended checkpoint (\texttt{BLEURT-20})\footnote{\url{https://github.com/google-research/bleurt}} as our baseline.

\textbf{NLI Based Metrics} methods also vary in their granularity.
We use a RoBERTa-large \citep{liu2019roberta} model finetuned\footnote{\url{https://huggingface.co/roberta-large-mnli}} on MultiNLI \citep{williams-etal-2018-broad} as a baseline for document-level evaluation, where the model evaluates a \textit{candidate} against the entire \textit{context}. Our baselines also include the DAE \citep{goyal-durrett-2020-evaluating} metric, which decomposes text at the level of dependency arcs. For sentence-level baseline, we use SummaC-ZeroShot and SummaC-Conv introduced in the SummaC Benchmark \citep{laban-etal-2022-summac} and FactCC \citep{kryscinski-etal-2020-evaluating} which is trained on synthetic data.

\textbf{Miscellaneous}\quad Besides the above metrics, we also use competitive metrics including UniEval \citep{Zhong2022TowardsAU}, CTC \citep{deng-etal-2021-compression}, BARTScore \citep{yuan2021bartscore} and BLANC \citep{vasilyev-etal-2020-fill} as baselines. 

UniEval is a unified multi-dimensional metric, capable of evaluating different aspects of text generation. We use the \texttt{Consistency} variant as the baseline. \citet{deng-etal-2021-compression} propose CTC, which is based on token-level information alignment. We use its discriminative variant trained on synthetic CNN/DailyMail \citep{see-etal-2017-get} (\texttt{D-CNNDM}) as our baseline. For BARTScore, we use the pretrained \texttt{BART-Large-CNN}\footnote{\url{https://github.com/neulab/BARTScore}} checkpoint.

\textbf{LLM-Based Metrics}\quad Concurrent work proposes to utilize LLMs for NLG evaluation. GPTScore uses the log probability of an LLM generating the target text conditioned on the prompt as the metric score \citep{fu-etal-2023-gptscore}. G-EVAL first augments its prompts with chain-of-thoughts and then evaluates texts by form-filling \citep{liu-etal-2023-g-eval}. \citet{gao-etal-2023-human} uses ChatGPT in place of human annotators in four popular human evaluation setups (\texttt{ChatGPT} in Table~\ref{tab:spearman-llm}). As we directly compare with correlation coefficients reported by \citet{fu-etal-2023-gptscore,liu-etal-2023-g-eval,gao-etal-2023-human}, results on some datasets are not available.

\begin{table*}[htbp]
\centering \footnotesize
\begin{tabular}{@{}l|lccccccc|c@{}}
\toprule
Type &
Metric &
XSF &
SE &
Q-X &
Q-C &
FRK-X &
FRK-C &
SSum &
\textbf{AVG} \\ \midrule
&
FEQA &
1.3 &
-2.9 &
-7.3 &
-3.9 &
3.0 &
-0.4 &
2.7 &
-1.0 \\
&
QuestEval &
\cellcolor[HTML]{E0F1CB}41.9 &
29.7 &
11.7 &
36.3 &
19.5 &
46.5 &
0.4 &
26.6 \\
\multirow{-3}{*}{QA} &
QAFactEval &
30.3 &
\cellcolor[HTML]{E0F1CB}61.6 &
44.2 &
68.4 &
32.1 &
\cellcolor[HTML]{E0F1CB}64.6 &
\cellcolor[HTML]{E0F1CB}38.9 &
\cellcolor[HTML]{E0F1CB}48.6 \\ \midrule
&
ROUGE-1 &
36.1 &
41.1 &
15.7 &
58.2 &
6.8 &
37.1 &
16.7 &
30.3 \\
&
ROUGE-2 &
27.6 &
40.9 &
14.4 &
59.2 &
4.9 &
38.7 &
19.1 &
29.3 \\
&
ROUGE-L &
30.6 &
42.3 &
12.5 &
58.2 &
8.0 &
37.7 &
17.4 &
29.5 \\
&
BLEU &
18.9 &
41.5 &
10.9 &
64.9 &
8.7 &
36.6 &
16.2 &
28.2 \\
&
BERTScore &
13.0 &
33.1 &
-10.6 &
51.7 &
13.0 &
51.7 &
10.9 &
23.3 \\
&
NER-Overlap &
21.9 &
24.9 &
31.2 &
0.3 &
11.4 &
30.1 &
16.7 &
19.5 \\
\multirow{-7}{*}{\begin{tabular}[c]{@{}l@{}}Similarity\\ Matching\end{tabular}} &
SimCSE &
30.9 &
28.5 &
11.9 &
48.6 &
13.5 &
34.5 &
10.7 &
25.5 \\ \midrule
Regression &
BLEURT &
38.7 &
23.8 &
13.2 &
45.2 &
15.6 &
37.5 &
8.1 &
26.0 \\ \midrule
&
MNLI &
15.8 &
-1.8 &
6.1 &
-11.0 &
19.7 &
-2.2 &
28.0 &
7.8 \\
&
DAE &
\cellcolor[HTML]{92D050}42.5 &
41.5 &
37.5 &
42.7 &
32.9 &
40.5 &
18.6 &
36.6 \\
&
SummaC-ZS &
6.4 &
50.1 &
43.7 &
56.1 &
14.7 &
53.7 &
13.7 &
34.0 \\
\multirow{-4}{*}{NLI} &
SummaC-CONV &
10.2 &
50.3 &
36.4 &
63.6 &
17.6 &
58.7 &
12.4 &
35.6 \\ \midrule
&
UniEval &
23.9 &
57.8 &
45.5 &
66.7 &
27.2 &
58.3 &
23.2 &
43.2 \\
&
CTC &
27.2 &
54.7 &
30.6 &
64.5 &
20.0 &
54.5 &
16.9 &
38.3 \\
&
BARTScore &
29.3 &
35.5 &
16.3 &
71.5 &
23.7 &
51.9 &
15.0 &
34.7 \\
&
FactCC &
4.9 &
34.8 &
28.8 &
38.6 &
8.3 &
34.8 &
-4.4 &
20.8 \\
\multirow{-5}{*}{Misc} &
BLANC &
8.3 &
21.3 &
1.8 &
25.7 &
6.4 &
34.3 &
8.3 &
15.2 \\ \midrule
&
\textbf{\alignmodel-base} &
38.2 &
61.1 &
\cellcolor[HTML]{E0F1CB}49.5 &
\cellcolor[HTML]{E0F1CB}72.3 &
\cellcolor[HTML]{E0F1CB}33.2 &
60.0 &
23.9 &
48.3 \\
\multirow{-2}{*}{Ours} &
\textbf{\alignmodel-large} &
31.1 &
\cellcolor[HTML]{92D050}66.3 &
\cellcolor[HTML]{92D050}52.7 &
\cellcolor[HTML]{92D050}78.1 &
\cellcolor[HTML]{92D050}38.3 &
\cellcolor[HTML]{92D050}67.7 &
\cellcolor[HTML]{92D050}44.6 &
\cellcolor[HTML]{92D050}54.1 \\ \bottomrule
\end{tabular}
\caption{Instance-level Pearson correlation coefficients on human annotated factual consistency datasets. The average performance of each metric is in column \textbf{AVG}. The color format is the same as in Table \ref{tab:summac-auc}. The full names of the datasets are listed in Table \ref{tab:other-dataset-names}.}
\label{tab:pearson}
\end{table*}

\subsection{Results}\label{sec:results}
\subsubsection{Results on SummaC Benchmark}
We report AUC-ROC on the test set of the SummaC Benchmark in Table \ref{tab:summac-auc}. A higher AUC-ROC score indicates the metric is better at detecting factual consistency errors. Our \alignmodel-large achieves the best average performance on the SummaC benchmark, scoring the highest in 4 out of 6 datasets.
We also present the balanced accuracy in Appendix (Table \ref{tab:summac-bacc}), where \alignmodel-large also establishes new state-of-the-art results. 

\subsubsection{Results on TRUE Benchmark} 
The results on the TRUE benchmark are shown in Table \ref{tab:true-benchmark}, where \alignmodel-large gets the highest average AUC-ROC score. It outperforms baselines on 7 out of 11 tasks while staying competitive on the rest. 
For a fair comparison, we also report the average AUC-ROC (denoted as \textbf{AVG-ZS}) excluding datasets that the alignment function is trained on (PAWS, VitaminC and FEVER). The performance of \alignmodel remains to be on top, outperforming strong baselines like QAFactEval, UniEval, and SummaC-CONV. This demonstrates \alignmodel generalizes well to unseen data (e.g., DialFact dataset in the dialogue domain).

\begin{figure}[t]
\centering
\includegraphics[width=0.95\columnwidth]{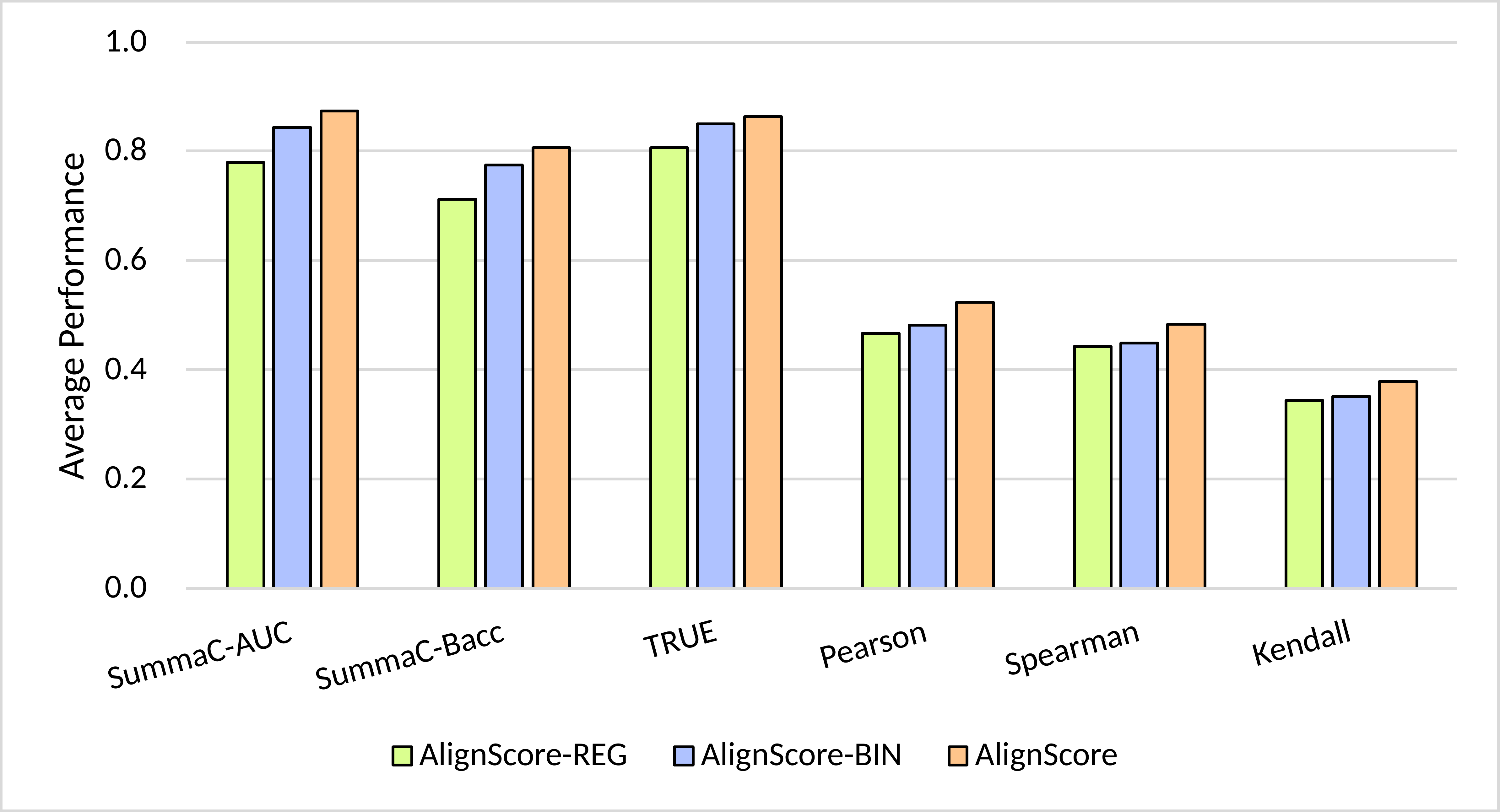}
\caption{The performance of \alignmodel-base using different classification heads. \alignmodel-REG and \alignmodel-BIN indicate the regression head and the binary classification head, respectively. \alignmodel is our proposed setting (see Section \ref{sec:consistency-metric}).}
\label{fig:compare-heads}
\end{figure}

\subsubsection{Results on Other Datasets}
We present Pearson correlation coefficients of various metrics on other factual consistency datasets in Table \ref{tab:pearson}. We also report Spearman correlation and Kendall's tau coefficients in Appendix (Table \ref{tab:spearman} and \ref{tab:kendall}). The \alignmodel-large metric outperforms previous metrics in terms of overall performance, including the competitive QAFactEval and UniEval metrics, dominating 6 out of 7 datasets.
We note that DAE and QuestEval perform better on XSumFaith dataset. Similar to \citet{fabbri-etal-2022-qafacteval}, we speculate it is because the relatedness between the token-level annotation of XSumFaith and the fine-grained metrics.

\begin{table}[t]
\centering
\resizebox{\columnwidth}{!}{%
\begin{tabular}{@{}llccc@{}}
\toprule
\multirow{2}{*}{Metric}    & \multirow{2}{*}{Backbone} & \multicolumn{3}{c}{Datasets}                  \\ \cmidrule(l){3-5} 
                           &                           & SE            & Q-X           & Q-C           \\ \midrule
G-EVAL-3.5                 & GPT3.5-d03                & 38.6          & 40.6          & 51.6          \\
G-EVAL-4                   & GPT4                      & \textbf{50.7} & 53.7          & 68.5          \\
GPTScore                   & GPT3.5-d03                & 47.5          & /             & /             \\
ChatGPT& GPT3.5-turbo             & 43.3          & /             & /             \\ \midrule
\textbf{\alignmodel-base}  & RoBERTa (125M)   & 43.4          & 51.9          & 69.0          \\
\textbf{\alignmodel-large} & RoBERTa (355M)   & 46.6          & \textbf{57.2} & \textbf{73.9} \\ \bottomrule
\end{tabular}%
}
\vspace{-6pt}
\caption{The Spearman correlation coefficients of \alignmodel and LLM-based metrics on SummEval (SE), QAGS-XSum (Q-X) and QAGS-CNNDM (Q-C). The best models are shown in \textbf{bold}. The results of G-EVAL, GPTScore and ChatGPT are from \citet{liu-etal-2023-g-eval}, \citet{fu-etal-2023-gptscore}, and \citet{gao-etal-2023-human}.}
\label{tab:spearman-llm}
\end{table}

We also compare our metric with LLM-based metrics in Table \ref{tab:spearman-llm}. Result shows \alignmodel has comparable performance with LLM-based metrics on SummEval. And it outperforms LLM-based metrics on QAGS-XSum and QAGS-CNNDM, showing the capability and efficiency of our proposed metric.

\subsection{Ablation Study}
To understand 1) which classification head is more suitable for factual consistency evaluation, 2) which splitting method is more effective, and 3) which upstream NLP task contributes the most to the superior performance of \alignmodel, we conduct 3 ablation studies. The experiments in this section are all based on \alignmodel-base.

\begin{figure}
\centering
\includegraphics[width=0.95\columnwidth]{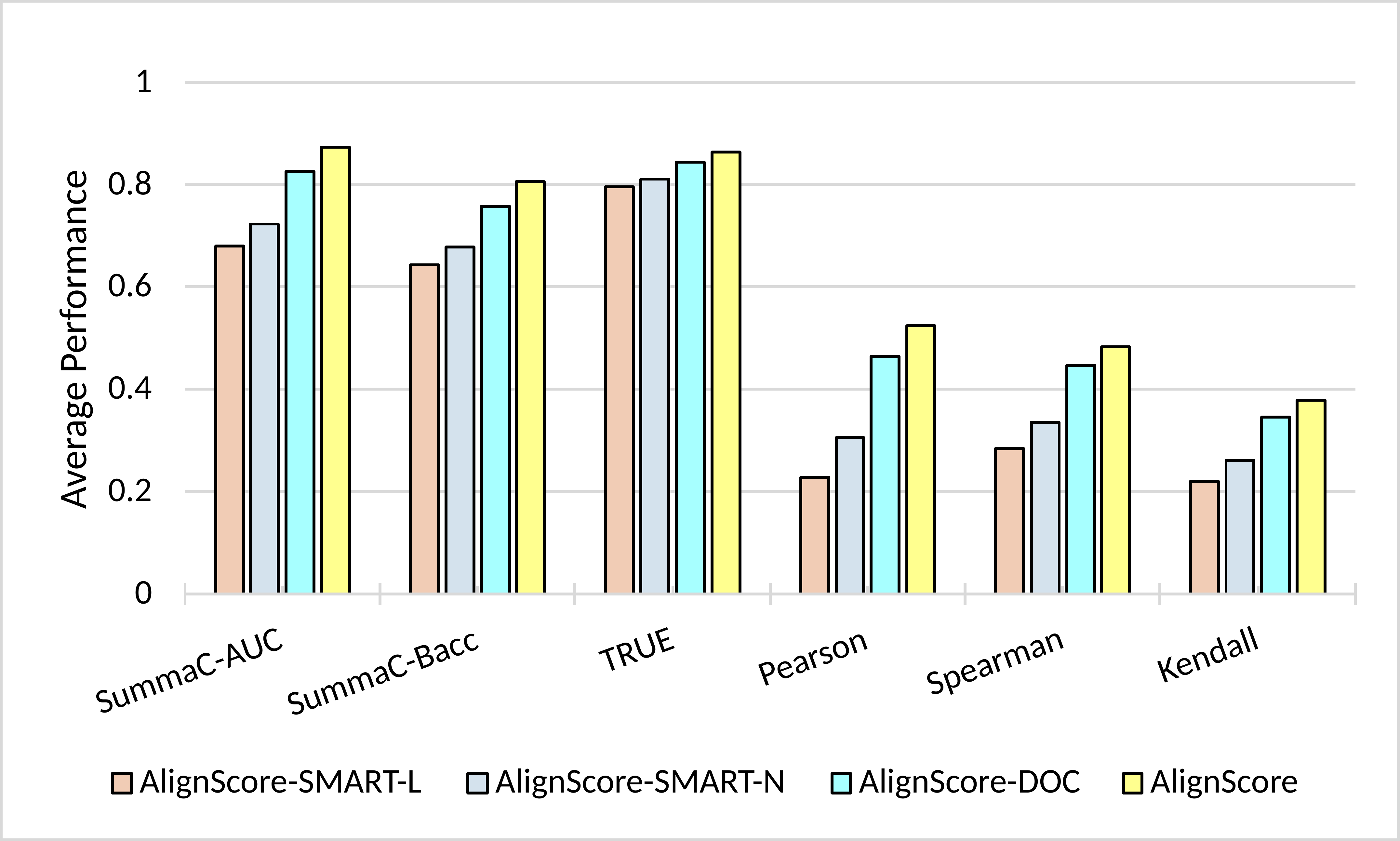}
\caption{The performance of \alignmodel-base using different splitting methods. \alignmodel-SMART-L and \alignmodel-SMART-N represent the SMART-L and SMART-N splitting methods, respectively. \alignmodel-DOC means no splitting (i.e. inputs are directly fed to the model). \alignmodel is our proposed splitting method (see Section \ref{sec:consistency-metric}).}
\label{fig:compare-split}
\end{figure}

\paragraph{Classification Head}\label{sec:alation-splitting}
We keep the same splitting method as in Section \ref{sec:consistency-metric} and change the heads that generate alignment scores. We first use the regression head (\alignmodel-base-REG) and the binary classification head (\alignmodel-base-BIN). Then, we compare these two heads with our proposed \alignmodel-base, which adopts the 3-way classification head. We present the results in Figure~\ref{fig:compare-heads}, which shows the 3-way classification head consistently performs better than the regression head and the binary classification head.

\paragraph{Splitting Method}
Then, we keep the 3-way classification head and change the splitting method. Following \citet{amplayo2022smart}, we implement SMART-L and SMART-N, and use our alignment model as the sentence matching function. SMART-L uses sentence-level evaluation and aggregates the alignment scores through a soft version of Longest Common Subsequence (LCS), while SMART-N aggregates using greedy matching between N-sentences. In our experiments, we set N=1. We also implement \alignmodel without any splitting (denoted as \alignmodel-base-DOC) where the inputs are directly fed into the model. The result in Figure \ref{fig:compare-split} shows that our chunk level splitting method performs best compared to the other 3 methods. It demonstrates that our splitting method helps \alignmodel capture salient information from long contexts.

\paragraph{Upstream NLP Task}\label{sec:ablation-task}
\begin{figure}
    \centering
    \includegraphics[width=1.0\columnwidth]{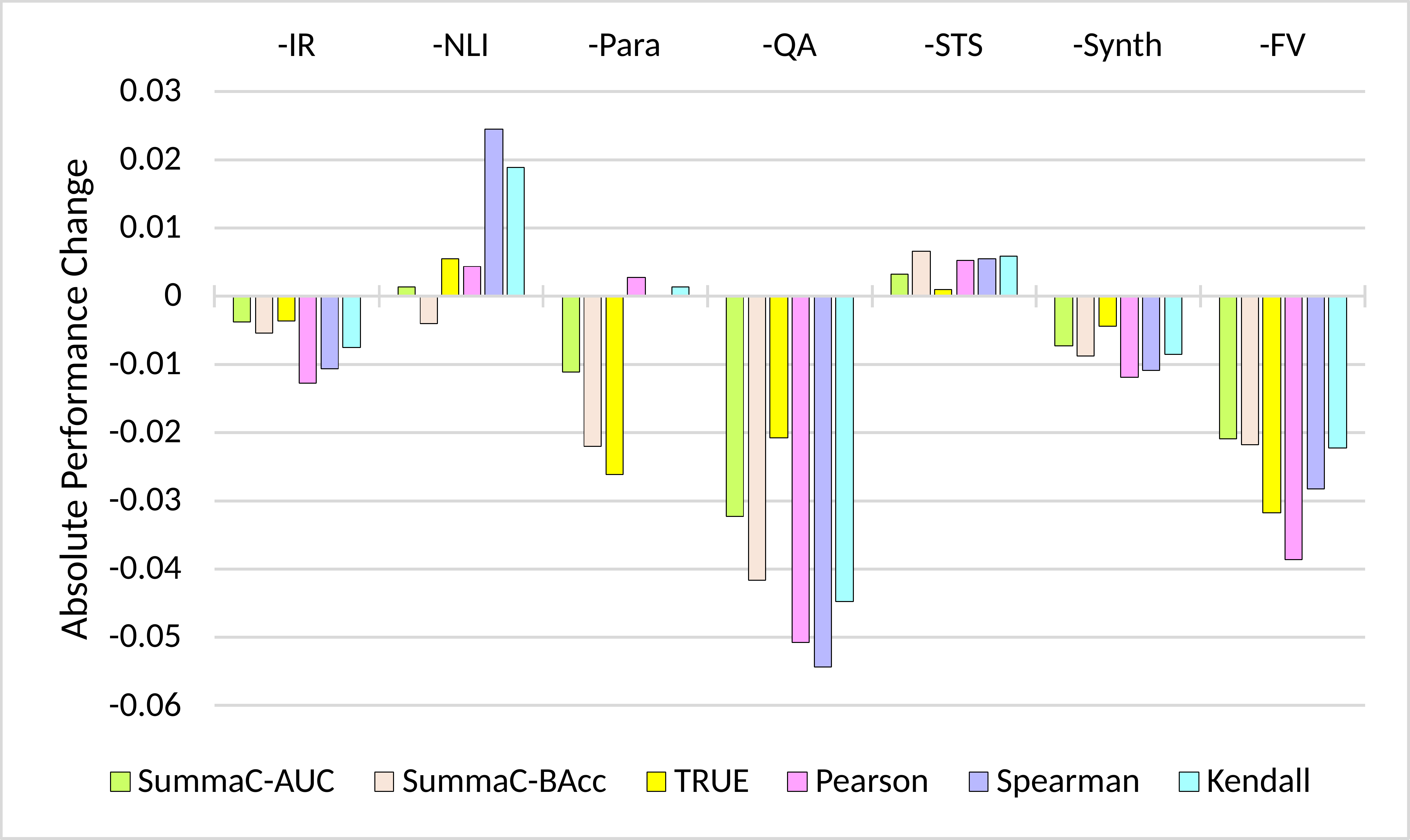}
    \caption{The absolute performance change of deducting one task when training alignment model. \textbf{-X} indicates the \textbf{X} task is removed from the alignment training. }
    \label{fig:ablation-pretrain-task}
\end{figure}
We study the contribution of each upstream NLP task by excluding one task at a time to train the alignment model. The results are shown in Figure \ref{fig:ablation-pretrain-task}. When the QA task is removed, the performance of the metric is the worst, indicating QA datasets make the biggest contribution to metric performance. Similarly, fact verification task has the second largest contribution. Surprisingly, with the removal of the NLI task, the model performs better on a majority of benchmarks, showing the NLI task plays a negative role in the training. We speculate that it is because 1) premises and hypothesises in NLI datasets are generally shorter, which differs from most factual consistency benchmarks and datasets, 2) other NLP tasks have larger-scale and higher quality datasets.

\section{Conclusion}
We propose \alignmodel, a holistic factual consistency metric based on a unified alignment function.
To learn the alignment function, we adapt 7 well established language understanding tasks into a unified alignment task
, resulting in 4.7M diverse training samples. 
Experiments show \alignmodel achieves state-of-the-art performance on SummaC and TRUE Benchmark, has higher correlation with human judgements than competing metrics, and generalizes well to unseen data.


\section*{Limitations}
\textit{Interpretability.}
Although \alignmodel shows high correlation with human judgments, it is hard to interpret the reasoning behind its predictions. Therefore, an interesting future research direction is to develop interpretable factual consistency metrics that can accurately identify words or spans in the input that contain factual consistency errors and (or) produce human readable explanations justifying its predictions.

\textit{Synthetic data.}
Our alignment training data contains datasets augmented with synthetic data. While ablation studies show that synthetic data helps improve metric performance, our rule-based method for generating synthetic data could generate noisy data that may not accurately model the error types and distributions produced by real world generative systems. Thus, analyzing the quality of synthetic data and developing more effective ways to generate synthetic data is an interesting research topic.

\textit{Language coverage.}
While we show \alignmodel generalize well to unseen data, it only covers a single language, English. Undoubtedly, factual consistency evaluation is also important for more resource-constrained languages or in a multilingual setting. Consequently, future research could focus on extending the Align metric to multiple languages, including resource-constrained languages.

\section*{Ethics Statement}
\alignmodel is intended as an automatic metric to be used in NLP research. While it has state-of-the-art performance, it can produce false positives and false negatives, and may not be appropriate for applications other than its intended use. As it is trained on publicly available datasets, the metric might be affected by biases inherent to those datasets.

\bibliography{anthology,custom}
\bibliographystyle{acl_natbib}

\appendix
\section{Implementation Details}

\subsection{Unifying Language Understanding Tasks}\label{sec:training-data}

We adapt datasets from 7 NLP tasks into the information alignment format. An overview of our unified training sets is shown in Table~\ref{tab:training-data}.




Tasks that cleanly fit into the form of the alignment problem, including NLI, fact verification, and paraphrase datasets are adapted by mapping the original labels into either binary or 3-way classification alignment labels. Next, we discuss how we adapt semantic textual similarity (STS), QA, and information retrieval (IR) tasks.

\paragraph{STS} STS datasets contain pairs of sentences labeled with semantic similarity scores. We use STS datasets in the regression task by normalizing the score to between 0 and 1.

\paragraph{QA} A QA sample consists of a context paragraph, a question, and a ground truth answer. One can derive the ground truth answer given the context and the question. To convert QA samples into a format suitable for binary classification, we use a pretrained sequence-to-sequence model to convert question-answer pairs into declarative sentences \citep{song-qa2d,demszky2018transforming}. Sentences generated from ground truth answers form \textsc{aligned} pairs with corresponding contexts, while sentences generated from wrong options form \textsc{not-aligned} samples. For samples with unanswerable questions, we first use a QA model\footnote{\url{https://huggingface.co/valhalla/t5-base-qa-qg-hl}}
to generate wrong answers, and then turn them into \textsc{not-aligned} samples using the above method.

See Section~\ref{sec:qa-samples} for converted samples.

\paragraph{IR} A sample in an information retrieval dataset consists of a query-answer pair and a list of passages, some of which can be used to answer the query. Similar to QA datasets, we adapt information retrieval datasets for binary classification by converting query-answer pairs into declarative sentences and then pairing them with passages. If a passage can be used to answer the corresponding query, we consider the sample to have \textsc{aligned} label. Otherwise it is assigned \textsc{not-aligned}.

\subsection{Synthetic Data}\label{sec:synthetic-data}
We further augment our training set with synthetic data based on the WikiText-103 corpus \citep{merity2017pointer} and the WikiHow summarization dataset \citep{koupaee2018wikihow}. 

To generate \textsc{aligned} samples, we create a paraphrase of each sentence in WikiText-103 through back translation using a neural machine translation model \citep{junczys-dowmunt-etal-2018-marian}. For the WikiHow dataset, we use source documents as text \(\boldsymbol{a}\), and the ground truth summaries together with extractive summaries generated by an extractive summarizer \citep{barrios2016variations} as text \(\boldsymbol{b}\) to form \textsc{aligned} samples.

Inspired by recent work in creating factually inconsistent samples \citep{deng-etal-2021-compression, kryscinski-etal-2020-evaluating}, we randomly mask 25\% of the tokens in text \(\boldsymbol{b}\) from the \textsc{aligned} samples and infill with a masked language modeling model \cite{sanh2019distilbert}. The resulting sentences are semantically different from the originals and are used in \textsc{not-aligned} samples.

\subsection{Training the Alignment Model} \label{sec:training-details}
We use the \texttt{Transformers}\footnote{\url{https://huggingface.co/docs/transformers/index}} library to implement the proposed model, and the PyTorch Lightning framework to train our model. 

The alignment model is optimized with AdamW \citep{loshchilov2018decoupled}. The learning rate is first warmed up to a peak of 1e-5, and then linearly decayed. The hyperparameters used to train \alignmodel-base and \alignmodel-large are shown in Table \ref{tab:hyperparams}.

We don't split the context and claims into chunks in the training for simplicity. 

\begin{table}[htbp]
\centering 
\resizebox{\columnwidth}{!}{%
\begin{tabular}{@{}lcc@{}}
\toprule
\textbf{Hyperparameter} & \textbf{\alignmodel-base} & \textbf{\alignmodel-large} \\ \midrule
Base Model              & RoBERTa-base        & RoBERTa-large        \\
Parameters              & 125M                & 355M                 \\
Batch Size              & 32                  & 32                   \\
Epochs                  & 3                   & 3                    \\
Optimizer               & AdamW               & AdamW                \\
Learning Rate           & 1e-5                & 1e-5                 \\
Weight Decay            & 0.1                 & 0.1                  \\
Adam $\epsilon$         & 1e-6                & 1e-6                 \\
Warmup Ratio            & 0.06                & 0.06                 \\
Random Seed             & 2022                & 2022                 \\
GPU                     & 2$\times$3090              & 4$\times$A5000              \\
GPU Hour                & 100h                & 532h           \\ \bottomrule
\end{tabular}
}
\caption{The hyperparameters used to train the alignment model.}
\label{tab:hyperparams}
\end{table}

\subsection{Cleaning Evaluation Datasets}
Certain datasets we use for evaluation contain artifacts that could hurt model performance. Notable issues include claims having escape sequences (\texttt{-LRB-} and \texttt{-RRB-} instead of parentheses) and being uncased (all lower case) while contexts do not have escape sequences and are cased.

We use rule-based methods to remove these artifacts. Specifically, we replace escape sequences in claims with the original characters, capitalize the first letter of the first word in a sentence, and for words that appear in contexts, we fix their capitalization in the corresponding claims according to their occurrences in the contexts.
\subsection{Computing Correlations}
We first split the inputs to sentences with \texttt{NLTK} sentenizer. Then \alignmodel computes the instance-level factual consistency score as stated in Section \ref{sec:consistency-metric}. We use \texttt{scipy} to compute Pearson correlation, Spearman correlation and Kendall's tau correlation.

\begin{table}[htbp]
\centering \footnotesize
\begin{tabular}{@{}ll@{}}
\toprule
\textbf{Dataset}    & \multicolumn{1}{l}{\textbf{Abbreviation}} \\ \midrule
SummEval   & SE                               \\
PAWS       & PAWS                             \\
Q2         & Q2                               \\
VitaminC   & VitC                             \\
FEVER      & FVR                              \\
FRANK      & FRK                              \\
DialFact   & DF                               \\
MNBM       & MNBM                             \\
QAGS-CNNDM & Q-C                              \\
QAGS-XSum  & Q-X                              \\
BEGIN      & BEGIN                            \\ \bottomrule
\end{tabular}
\caption{The abbreviations of each dataset in TRUE benchmark.}
\label{tab:true-abbr}
\end{table}

\begin{table}[htbp]
\centering \footnotesize
\begin{tabular}{@{}ll@{}}
\toprule
\textbf{Dataset} & \multicolumn{1}{l}{\textbf{Abbreviation}} \\ \midrule
XSumFaith        & XSF                                       \\
SummEval         & SE                                        \\
QAGS-Xsum        & Q-X                                       \\
QAGS-CNNDM       & Q-C                                       \\
FRANK-XSum       & FRK-X                                     \\
FRANK-CNNDM      & FRK-C                                     \\
SamSum           & SSum                                      \\ \bottomrule
\end{tabular}
\caption{The abbreviations of each dataset in Table \ref{tab:pearson}/\ref{tab:spearman}/\ref{tab:kendall}.}
\label{tab:other-dataset-names}
\end{table}

\section{Additional Experiment Details/Results}
\label{sec:additional-exp-result}

\subsection{SummaC Benchmark}
SummaC benchmark consists of 6 summarization datasets: CogenSum \citep{falke-etal-2019-ranking}, XSumFaith \citep{maynez-etal-2020-faithfulness}, Polytope \citep{huang-etal-2020-achieved}, FactCC \citep{kryscinski-etal-2020-evaluating}, SummEval \citep{fabbri-etal-2021-summeval} and FRANK \citep{pagnoni-etal-2021-understanding}. The datasets are standardized by binarizing each labels. Metrics are evaluated as classifiers on SummaC benchmark.

The SummaC Benchmark considers samples in PolyTope with \texttt{Addition}\footnote{Defined as: Unnecessary and irrelevant snippets from the source are included in the summary}, \texttt{Omission}\footnote{Defined as: Key point is missing from the output}, \texttt{Inaccuracy Intrinsic}\footnote{Defined as: Terms or concepts from the source are misrepresented and thus unfaithful.}, \texttt{Inaccuracy Extrinsic}\footnote{Defined as: The summary has content not presented in the source and factually incorrect} and \texttt{Positive-Negative Aspect}\footnote{Defined as: The output summary represents positive statements whereas the source segment is negative, and vice versa.} errors to be negative samples. However, \texttt{Addition} and \texttt{Omission} do not imply factual consistency errors. Thus, we only consider samples with \texttt{Inaccuracy Intrinsic}, \texttt{Inaccuracy Extrinsic} and \texttt{Positive-Negative Aspect} errors to be factually incorrect. The reported PolyTope result uses this definition of errors.


We also report balanced accuracy, which deals with imbalanced datasets, in Table \ref{tab:summac-bacc}. 

\begin{table*}[htbp]
\centering \footnotesize
\begin{tabular}{@{}c|lcccccc|c@{}}
\toprule
Type &
Metric &
CGS &
XSF &
PolyTope &
FactCC &
SummEval &
FRANK &
\textbf{AVG} \\ \midrule
&
FEQA &
51.9 &
49.5 &
53.7 &
46.6 &
51.4 &
41.4 &
49.1 \\
&
QuestEval &
53.1 &
57.6 &
69.3 &
66.8 &
69.8 &
77.7 &
65.7 \\
\multirow{-3}{*}{QA} &
QAFactEval &
50.6 &
61.2 &
60.2 &
73.8 &
54.9 &
74.9 &
62.6 \\ \midrule
&
ROUGE-1 &
61.1 &
62.4 &
74.4 &
68.0 &
80.0 &
79.1 &
70.8 \\
&
ROUGE-2 &
61.2 &
62.2 &
75.1 &
67.8 &
78.8 &
78.8 &
70.7 \\
&
ROUGE-L &
61.5 &
57.4 &
74.0 &
67.7 &
79.7 &
78.8 &
69.8 \\
&
BLEU &
64.2 &
55.2 &
78.3 &
67.0 &
77.6 &
79.3 &
70.3 \\
&
BERTScore &
52.7 &
49.0 &
76.9 &
65.3 &
72.7 &
78.5 &
65.8 \\
&
NER-Overlap &
51.1 &
64.9 &
72.1 &
49.8 &
56.6 &
68.1 &
60.4 \\
\multirow{-7}{*}{\begin{tabular}[c]{@{}c@{}}Similarity\\ Matching\end{tabular}} &
SimCSE &
54.4 &
57.3 &
68.9 &
57.3 &
71.3 &
68.5 &
62.9 \\ \midrule
Regression &
BLEURT &
57.7 &
58.7 &
69.0 &
56.2 &
63.7 &
74.9 &
63.4 \\ \midrule
&
MNLI &
46.0 &
48.7 &
46.3 &
52.2 &
50.7 &
55.2 &
49.8 \\
&
DAE &
52.4 &
\cellcolor[HTML]{92D050}76.7 &
72.8 &
54.2 &
66.1 &
78.9 &
66.8 \\
&
SummaC-ZS &
62.6 &
57.8 &
81.0 &
82.8 &
77.8 &
78.1 &
73.4 \\
\multirow{-4}{*}{NLI} &
SummaC-CONV &
59.8 &
66.4 &
73.7 &
\cellcolor[HTML]{E0F1CB}89.2 &
79.8 &
81.0 &
75.0 \\ \midrule
&
UniEval &
\cellcolor[HTML]{E0F1CB}77.1 &
61.2 &
\cellcolor[HTML]{E0F1CB}85.3 &
84.7 &
79.4 &
80.9 &
78.1 \\
&
CTC &
69.1 &
61.7 &
82.1 &
77.6 &
78.4 &
80.5 &
74.9 \\
&
BARTScore &
56.9 &
58.7 &
84.6 &
73.3 &
79.6 &
78.3 &
71.9 \\
&
FactCC &
64.9 &
55.1 &
78.5 &
72.7 &
71.8 &
69.8 &
68.8 \\
\multirow{-5}{*}{Misc} &
BLANC &
49.8 &
52.0 &
66.3 &
55.7 &
58.3 &
78.4 &
60.1 \\ \midrule
&
\textbf{\alignmodel-base} &
\cellcolor[HTML]{92D050}77.8 &
\cellcolor[HTML]{E0F1CB}72.2 &
78.9 &
87.4 &
\cellcolor[HTML]{92D050}83.7 &
\cellcolor[HTML]{E0F1CB}83.6 &
\cellcolor[HTML]{E0F1CB}80.6 \\
\multirow{-2}{*}{Ours} &
\textbf{\alignmodel-large} &
75.0 &
70.0 &
\cellcolor[HTML]{92D050}88.0 &
\cellcolor[HTML]{92D050}89.2 &
\cellcolor[HTML]{E0F1CB}83.4 &
\cellcolor[HTML]{92D050}86.3 &
\cellcolor[HTML]{92D050}82.0 \\ \bottomrule
\end{tabular}
\caption{Balanced accuracy of various metrics on SummaC benchmark. We compute the averaged performance of each metric in the last column \textbf{AVG}. The color format follows Table \ref{tab:summac-auc}.}
\label{tab:summac-bacc}
\end{table*}

\subsection{TRUE Benchmark}
TRUE benchmark is for evaluating factual consistency metrics in summarization, dialogue, fact-verification and paraphrasing tasks. There are totally 11 datasets in this benchmark: FRANK \citep{pagnoni-etal-2021-understanding}, SummEval \citep{fabbri-etal-2021-summeval}, MNBM \citep{maynez-etal-2020-faithfulness}, QAGS-CNNDM \citep{wang-etal-2020-asking}, QAGS-XSum \citep{wang-etal-2020-asking}, BEGIN \citep{dziri-etal-2022-evaluating}, Q$^2_\text{dataset}$ \citep{honovich-etal-2021-q2}, DialFact \citep{gupta-etal-2022-dialfact}, PAWS \citep{zhang-etal-2019-paws}, FEVER \citep{DBLP:conf/aaai/NieCB19, thorne-etal-2018-fever} and VitaminC \citep{schuster-etal-2021-get}. TRUE also treats factual consistency evaluation as a binary classification task and reports AUC-ROC.

The full names of the datasets in Table \ref{tab:true-benchmark} are listed in Table \ref{tab:true-abbr}.

\subsection{Other Datasets}

\begin{table*}[htbp]
\centering \footnotesize
\begin{tabular}{@{}c|lccccccc|c@{}}
\toprule
Type &
  Metric &
  XSF &
  SE &
  Q-X &
  Q-C &
  FRK-X &
  FRK-C &
  SSum &
  \textbf{AVG} \\ \midrule
 &
  FEQA &
  1.7 &
  0.2 &
  -6.5 &
  -7.2 &
  1.5 &
  -2.9 &
  0.0 &
  -1.9 \\
 &
  QuestEval &
  42.1 &
  26.3 &
  11.9 &
  30.8 &
  19.1 &
  40.5 &
  3.9 &
  25.0 \\
\multirow{-3}{*}{QA} &
  QAFactEval &
  31.9 &
  42.8 &
  44.1 &
  63.1 &
  25.5 &
  53.7 &
  \cellcolor[HTML]{E0F1CB}35.9 &
  42.4 \\ \midrule
 &
  ROUGE-1 &
  34.2 &
  38.1 &
  18.1 &
  53.6 &
  5.6 &
  35.2 &
  15.1 &
  28.6 \\
 &
  ROUGE-2 &
  26.8 &
  37.8 &
  17.7 &
  55.2 &
  2.8 &
  37.2 &
  17.5 &
  27.9 \\
 &
  ROUGE-L &
  28.9 &
  38.5 &
  16.5 &
  53.7 &
  8.2 &
  35.8 &
  16.3 &
  28.3 \\
 &
  BLEU &
  18.2 &
  34.7 &
  10.1 &
  55.4 &
  6.3 &
  34.0 &
  13.7 &
  24.6 \\
 &
  BERTScore &
  13.4 &
  31.5 &
  -8.9 &
  46.2 &
  12.7 &
  45.1 &
  13.1 &
  21.9 \\
 &
  NER-Overlap &
  23.9 &
  21.4 &
  31.2 &
  0.2 &
  11.3 &
  27.8 &
  16.7 &
  18.9 \\
\multirow{-7}{*}{\begin{tabular}[c]{@{}c@{}}Similarity\\ Matching\end{tabular}} &
  SimCSE &
  29.2 &
  26.4 &
  11.2 &
  47.2 &
  13.3 &
  31.3 &
  7.9 &
  23.8 \\ \midrule
Regression &
  BLEURT &
  37.0 &
  23.6 &
  12.4 &
  43.4 &
  13.9 &
  37.6 &
  6.7 &
  24.9 \\ \midrule
 &
  MNLI &
  7.0 &
  -6.6 &
  0.7 &
  -16.4 &
  11.7 &
  -5.5 &
  31.1 &
  3.1 \\
 &
  DAE &
  \cellcolor[HTML]{92D050}47.0 &
  36.2 &
  37.5 &
  37.1 &
  \cellcolor[HTML]{92D050}32.1 &
  36.9 &
  18.6 &
  35.1 \\
 &
  SummaC-ZS &
  5.7 &
  38.3 &
  43.7 &
  51.1 &
  12.8 &
  46.2 &
  15.1 &
  30.4 \\
\multirow{-4}{*}{NLI} &
  SummaC-CONV &
  21.7 &
  41.4 &
  45.0 &
  58.4 &
  11.0 &
  52.4 &
  9.8 &
  34.2 \\ \midrule
 &
  UniEval &
  25.3 &
  \cellcolor[HTML]{E0F1CB}44.3 &
  50.0 &
  67.6 &
  26.7 &
  54.0 &
  22.8 &
  41.5 \\
 &
  CTC &
  29.8 &
  41.7 &
  30.6 &
  57.3 &
  20.4 &
  49.4 &
  17.7 &
  35.3 \\
 &
  BARTScore &
  29.8 &
  39.1 &
  17.0 &
  68.1 &
  20.0 &
  53.3 &
  16.3 &
  34.8 \\
 &
  FactCC &
  6.8 &
  33.5 &
  28.8 &
  40.3 &
  7.9 &
  35.3 &
  -4.4 &
  21.2 \\
\multirow{-5}{*}{Misc} &
  BLANC &
  8.4 &
  19.0 &
  1.6 &
  22.2 &
  6.5 &
  34.2 &
  9.1 &
  14.4 \\ \midrule
 &
  \textbf{\alignmodel-base} &
  \cellcolor[HTML]{E0F1CB}43.8 &
  43.4 &
  \cellcolor[HTML]{E0F1CB}51.9 &
  \cellcolor[HTML]{E0F1CB}69.0 &
  28.0 &
  \cellcolor[HTML]{E0F1CB}54.7 &
  23.4 &
  \cellcolor[HTML]{E0F1CB}44.9 \\
\multirow{-2}{*}{Ours} &
  \textbf{\alignmodel-large} &
  33.3 &
  \cellcolor[HTML]{92D050}46.6 &
  \cellcolor[HTML]{92D050}57.2 &
  \cellcolor[HTML]{92D050}73.9 &
  \cellcolor[HTML]{E0F1CB}29.0 &
  \cellcolor[HTML]{92D050}60.9 &
  \cellcolor[HTML]{92D050}43.8 &
  \cellcolor[HTML]{92D050}49.3 \\ \bottomrule
\end{tabular}

\caption{Instance-level Spearman correlation coefficients on human annotated factual consistency datasets. The table format follows Table \ref{tab:pearson}.}
\label{tab:spearman}
\end{table*}

\begin{table*}[htbp]
\centering \footnotesize
\begin{tabular}{@{}c|lccccccc|c@{}}
\toprule
Type &
  Metric &
  XSF &
  SE &
  Q-X &
  Q-C &
  FRK-X &
  FRK-C &
  SSum &
  \textbf{AVG} \\ \midrule
 &
  FEQA &
  1.1 &
  0.2 &
  -5.3 &
  -5.7 &
  1.3 &
  -2.2 &
  0.0 &
  -1.5 \\
 &
  QuestEval &
  28.7 &
  20.8 &
  9.7 &
  23.9 &
  15.6 &
  31.1 &
  3.2 &
  19.0 \\
\multirow{-3}{*}{QA} &
  QAFactEval &
  23.2 &
  34.0 &
  36.2 &
  50.5 &
  22.4 &
  42.2 &
  \cellcolor[HTML]{E0F1CB}30.1 &
  34.1 \\ \midrule
 &
  ROUGE-1 &
  23.4 &
  30.3 &
  14.8 &
  42.9 &
  4.6 &
  26.8 &
  12.4 &
  22.2 \\
 &
  ROUGE-2 &
  18.4 &
  30.0 &
  14.5 &
  44.2 &
  2.3 &
  28.4 &
  14.5 &
  21.8 \\
 &
  ROUGE-L &
  19.6 &
  30.6 &
  13.6 &
  42.8 &
  6.7 &
  27.3 &
  13.3 &
  22.0 \\
 &
  BLEU &
  14.6 &
  27.5 &
  9.0 &
  44.7 &
  6.1 &
  25.9 &
  12.2 &
  20.0 \\
 &
  BERTScore &
  9.2 &
  24.9 &
  -7.3 &
  36.3 &
  10.4 &
  34.7 &
  10.7 &
  17.0 \\
 &
  NER-Overlap &
  19.6 &
  20.6 &
  31.2 &
  0.2 &
  11.3 &
  25.7 &
  16.7 &
  17.9 \\
\multirow{-7}{*}{\begin{tabular}[c]{@{}c@{}}Similarity\\ Matching\end{tabular}} &
  SimCSE &
  19.9 &
  20.9 &
  9.1 &
  36.7 &
  10.8 &
  23.8 &
  6.4 &
  18.2 \\ \midrule
Regression &
  BLEURT &
  25.3 &
  18.6 &
  10.1 &
  33.9 &
  11.4 &
  28.8 &
  5.5 &
  19.1 \\ \midrule
 &
  MNLI &
  4.7 &
  -5.2 &
  0.5 &
  -12.8 &
  9.5 &
  -4.2 &
  25.4 &
  2.6 \\
 &
  DAE &
  \cellcolor[HTML]{92D050}38.6 &
  34.8 &
  37.5 &
  34.7 &
  \cellcolor[HTML]{92D050}32.1 &
  34.1 &
  18.6 &
  32.9 \\
 &
  SummaC-ZS &
  3.9 &
  30.4 &
  35.8 &
  40.5 &
  10.5 &
  35.8 &
  12.3 &
  24.2 \\
\multirow{-4}{*}{NLI} &
  SummaC-CONV &
  15.0 &
  33.1 &
  36.8 &
  46.5 &
  9.0 &
  41.3 &
  8.0 &
  27.1 \\ \midrule
 &
  UniEval &
  17.0 &
  \cellcolor[HTML]{E0F1CB}35.3 &
  40.9 &
  54.4 &
  21.8 &
  42.4 &
  18.7 &
  32.9 \\
 &
  CTC &
  20.2 &
  33.2 &
  25.1 &
  45.7 &
  16.6 &
  38.2 &
  14.4 &
  27.6 \\
 &
  BARTScore &
  20.2 &
  31.0 &
  13.9 &
  \cellcolor[HTML]{E0F1CB}55.6 &
  16.3 &
  41.4 &
  13.3 &
  27.4 \\
 &
  FactCC &
  5.6 &
  32.2 &
  28.8 &
  37.7 &
  7.9 &
  32.6 &
  -4.4 &
  20.0 \\
\multirow{-5}{*}{Misc} &
  BLANC &
  5.6 &
  14.9 &
  1.3 &
  17.1 &
  5.3 &
  26.0 &
  7.5 &
  11.1 \\ \midrule
 &
  \textbf{\alignmodel-base} &
  \cellcolor[HTML]{E0F1CB}30.1 &
  34.7 &
  \cellcolor[HTML]{E0F1CB}42.5 &
  55.4 &
  22.9 &
  \cellcolor[HTML]{E0F1CB}42.9 &
  19.1 &
  \cellcolor[HTML]{E0F1CB}35.4 \\
\multirow{-2}{*}{Ours} &
  \textbf{\alignmodel-large} &
  22.7 &
  \cellcolor[HTML]{92D050}37.4 &
  \cellcolor[HTML]{92D050}46.8 &
  \cellcolor[HTML]{92D050}61.3 &
  \cellcolor[HTML]{E0F1CB}23.7 &
  \cellcolor[HTML]{92D050}48.5 &
  \cellcolor[HTML]{92D050}35.8 &
  \cellcolor[HTML]{92D050}39.5 \\ \bottomrule
\end{tabular}
\caption{Instance-level Kendall's tau correlation coefficients on human annotated factual consistency datasets. The table format follows Table \ref{tab:pearson}.}
\label{tab:kendall}
\end{table*}

In addition to the Pearson correlation reported in Table \ref{tab:pearson}, we also report the Spearman correlation and Kendall's tau correlation on 9 datasets in Table \ref{tab:spearman} and \ref{tab:kendall}, respectively. The full names of the abbreviations in Table \ref{tab:pearson}, Table \ref{tab:spearman} and Table \ref{tab:kendall} are listed in Table \ref{tab:other-dataset-names}.

\subsubsection{Why BLEU Metric Performs Relatively Well?}
We notice that the BLEU metric has comparable performance with some neural model based methods, which seems to contradict some previous findings. We attribute it to the case matching in the pre-processing, since BLEU is case sensitive.

\section{Sample Training Data}

\subsection{Converted QA Samples}\label{sec:qa-samples}

We show converted SQuAD v2 \citep{rajpurkar-etal-2018-know} samples below to illustrate the process of converting QA samples into the alignment format (discussed in Section~\ref{sec:training-data}). Concretely, questions and answers are combined into declarative claims using a sequence-to-sequence model \citep{song-qa2d,demszky2018transforming}.

\begin{description}[leftmargin=!,labelwidth=0.25in,itemsep=0in,topsep=0.3in]
  \item [Context:] The Times Literary Supplement (TLS) first appeared in 1902 as a supplement to The Times, becoming a separately paid-for weekly literature and society magazine in 1914. The Times and the TLS have continued to be co-owned, and as of 2012 the TLS is also published by News International and cooperates closely with The Times, with its online version hosted on The Times website, and its editorial offices based in Times House, Pennington Street, London.
  \item [Question:] The editorial offices of The Times Literary Supplement is based in what location in London?
  \item [Answer:] Times House, Pennington Street
  \item [Generated claim:] The editorial offices of The Times Literary Supplement is based in Times House, Pennington Street in London.
  \item [Label:] \textsc{\textsc{aligned}}
\end{description}

\begin{description}[leftmargin=!,labelwidth=0.25in,itemsep=0in,topsep=0.3in]
  \item [Context:] The 25,000 cotton growers in the United States of America are heavily subsidized at the rate of \$2 billion per year although China now provides the highest overall level of cotton sector support. The future of these subsidies is uncertain and has led to anticipatory expansion of cotton brokers' operations in Africa. Dunavant expanded in Africa by buying out local operations. This is only possible in former British colonies and Mozambique; former French colonies continue to maintain tight monopolies, inherited from their former colonialist masters, on cotton purchases at low fixed prices.
  \item [Question:] How many subsidized cotton growers are in the US?
  \item [Answer:] 25,000
  \item [Generated claim:] 25,000 subsidized cotton growers are in the US.
  \item [Label:] \textsc{\textsc{aligned}}
\end{description}

\begin{description}[leftmargin=!,labelwidth=0.25in,itemsep=0in,topsep=0.3in]
  \item [Context:] On October 28, 2015, IBM announced its acquisition of digital assets from The Weather Company—a holding company of Bain Capital, The Blackstone Group and NBCUniversal which owns The Weather Channel, including its weather data platforms (such as Weather Services International), websites (Weather.com and Weather Underground) and mobile apps. The acquisition seeks to use Watson for weather analytics and predictions. The acquisition does not include The Weather Channel itself, which will enter into a long-term licensing agreement with IBM for use of its data. The sale closed on January 29, 2016 
  \item [Question:] When did the sale of Weather Company assets close?
  \item [Answer:] January 29, 2016
  \item [Generated claim:] The sale of Weather Company assets closed on January 29, 2016.
  \item [Label:] \textsc{\textsc{aligned}}
\end{description}

\begin{description}[leftmargin=!,labelwidth=0.25in,itemsep=0in,topsep=0.3in]
  \item [Context:] The dipole component of the magnetic field at the magnetic equator of Neptune is about 14 microteslas (0.14 G). The dipole magnetic moment of Neptune is about 2.2 × 1017 T·m3 (14 \(\mu\)T·RN3, where RN is the radius of Neptune). Neptune's magnetic field has a complex geometry that includes relatively large contributions from non-dipolar components, including a strong quadrupole moment that may exceed the dipole moment in strength. By contrast, Earth, Jupiter and Saturn have only relatively small quadrupole moments, and their fields are less tilted from the polar axis. The large quadrupole moment of Neptune may be the result of offset from the planet's centre and geometrical constraints of the field's dynamo generator.
  \item [Question:] What is the dipole component of the magnetic field at the magnetic equator of neptune?
  \item [Answer:] 14 microteslas (0.14 G)
  \item [Generated claim:] The dipole component of the magnetic field at the magnetic equator of neptune is 14 microteslas (0.14 G).
  \item [Label:] \textsc{\textsc{aligned}}
\end{description}

\begin{description}[leftmargin=!,labelwidth=0.25in,itemsep=0in,topsep=0.3in]
  \item [Context:] Qing dynasty rule in Tibet began with their 1720 expedition to the country when they expelled the invading Dzungars. Amdo came under Qing control in 1724, and eastern Kham was incorporated into neighbouring Chinese provinces in 1728. Meanwhile, the Qing government sent resident commissioners called Ambans to Lhasa. In 1750 the Ambans and the majority of the Han Chinese and Manchus living in Lhasa were killed in a riot, and Qing troops arrived quickly and suppressed the rebels in the next year. Like the preceding Yuan dynasty, the Manchus of the Qing dynasty exerted military and administrative control of the region, while granting it a degree of political autonomy. The Qing commander publicly executed a number of supporters of the rebels and, as in 1723 and 1728, made changes in the political structure and drew up a formal organization plan. The Qing now restored the Dalai Lama as ruler, leading the governing council called Kashag, but elevated the role of Ambans to include more direct involvement in Tibetan internal affairs. At the same time the Qing took steps to counterbalance the power of the aristocracy by adding officials recruited from the clergy to key posts.
  \item [Question:] What did the Qing commander do in 1732 and 1728?
  \item [Answer:] \texttt{Unanswerable}
  \item [Generated claim:] The Qing commander publicly executed a number of supporters of the rebels in 1732 and 1728.
  \item [Label:] \textsc{\textsc{not-aligned}}
\end{description}

\end{document}